\pdfoutput=1

\documentclass[11pt]{article}

\usepackage[]{EACL2023}

\usepackage{times}
\usepackage{latexsym}

\usepackage[T1]{fontenc}

\usepackage[utf8]{inputenc}

\usepackage{microtype}

\usepackage{inconsolata}

\usepackage{amsmath, mathtools}
\usepackage[super]{nth}

\usepackage{multirow}
\usepackage{booktabs, tabularx}
\usepackage{amssymb}
\usepackage{subfig}
\usepackage{dblfloatfix}
\usepackage{footmisc}

\newcommand\blfootnote[1]{%
  \begingroup
  \renewcommand\thefootnote{}\footnote{#1}%
  \addtocounter{footnote}{-1}%
  \endgroup
}

\title{Guide the Learner: Controlling Product of Experts Debiasing Method Based on Token Attribution Similarities}

\author{
    Ali Modarressi$^{1\star}$ ~
    \textbf{Hossein Amirkhani$^{2}$}  ~ \textbf{Mohammad Taher Pilehvar$^{3}$} \\
    $^1$ Center for Information and Language Processing, LMU Munich \\
    $^2$ Computer and Information Technology Department, University
of Qom, Iran \\
    $^3$ Tehran Institute for Advanced Studies, Khatam University, Iran \\
    \texttt{amodaresi@cis.lmu.de}\\
    \texttt{amirkhani@qom.ac.ir}\\
    \texttt{mp792@cam.ac.uk}
}

\begin{document}
\maketitle
\begin{abstract}
Several proposals have been put forward in recent years for improving out-of-distribution (OOD) performance through mitigating dataset biases.
A popular workaround is to train a robust model by re-weighting training examples based on a secondary biased model.
Here, the underlying assumption is that the biased model resorts to shortcut features. Hence, those training examples that are correctly predicted by the biased model are flagged as being biased and are down-weighted during the training of the main model.
However, assessing the importance of an instance merely based on the  predictions of the biased model may be too naive.
It is possible that the prediction of the main model can be derived from another decision-making process that is distinct from the behavior of the biased model.
To circumvent this, we introduce a fine-tuning strategy that incorporates the similarity between the main and biased model attribution scores in a Product of Experts (PoE) loss function to further improve OOD performance. 
With experiments conducted on natural language inference and fact verification benchmarks, we show that our method improves OOD results while maintaining in-distribution (ID) performance.\blfootnote{$^\star$ Work done as a Master's student at Iran University of Science and Technology (IUST).}\footnote{Our code is freely available at: \href{https://github.com/amodaresi/Debias_w_Saliencies}{\texttt{https://github.com/
amodaresi/Debias\_w\_Saliencies}}}
\end{abstract}

\section{Introduction}
Overfitting to the training data is a big obstacle in learning patterns that generalize to unseen data. Traditionally, this is diagnosed by monitoring the performance of a trained model on an in-distribution (ID) test set. However, a bigger challenge is when both the training and test data have the same non-generalizable patterns, emerged as spurious correlations between input features and output labels~\cite{gardner2021competency}. For instance, in the natural language inference (NLI) task, it is shown that the occurrence of some task-neutral words, like a negation in hypothesis, is highly correlated with a specific class~\citep{gururangan2018annotation}. 
While high-capacity models can learn a generalized distribution of labels from the inputs, they are prone to spurious patterns, also known as dataset biases \cite{clark2019don, he-etal-2019-unlearn}. 
A model could exploit these biases during fine-tuning, leading to a model that achieve high ID performance, while it is highly fragile in out-of-distribution (OOD) settings~\citep{schuster-etal-2019-towards,mccoy2020right}.

Besides trying to prevent these non-generalizable artifacts from entering the dataset~\cite{liu2022wanli}, it is reasonable to seek for more robust learning methods. 
This has been the basis for a multitude of research works that encourage models to rely on truly generalizable patterns. 
Most of these methods are based on the assumption that the learning method will inevitably exploit biases if they are present in a training example~\citep{clark2019don,sanh2020learning,karimi2020end,utama2020towards,ghaddar2021end}. Therefore, they discourage the main model from paying much attention to the examples which are correctly classified by a biased model. 
Recently, it is shown that this assumption is questionable in the way that for a significant number of cases, the main model does not follow the biased model in treating biased examples~\citep{amirkhani2021don}. 
Therefore, depriving the training algorithm from the examples which are detected to be biased is a waste of training data. 

In this paper, we propose an alternative way to discard biased examples. Instead of considering the mere prediction of the biased model, we monitor the way the model processes each example by computing its attribution scores over the input tokens.
With the resulting scores, we adjust the proportion of the loss function that is a cross-entropy loss (CE) versus a Product of Experts loss (PoE). 
If the attribution scores are similar between the main and biased models, the loss becomes a PoE loss where a correct prediction from the biased model down-weights the contribution of the corresponding example. 
In contrast, dissimilarity between the scores suggests a different behaviour from the biased model and leads to a CE loss that only considers the main model's prediction. 
Experiments on natural language inference and fact verification demonstrate that our method significantly outperforms previous approaches in terms of OOD performance while preserving its ID performance.


\section{Methodology}
In this section we explain our debiasing solution for a given classification task with dataset $\mathcal{D}\{\boldsymbol{x_i},y_i\}_{i=1}^N$. For the $i^{\text{th}}$ training example, we denote the input sequence of tokens as $\boldsymbol{x_i}$ and the gold label as $y_i \in \{1,2,...,Y\}$ where $Y$ is the number of classes. 
Specifically, the goal is to train a model---we denote it as the \emph{main model}---on the training dataset ($\mathcal{D}$) so that it can also perform well on OOD datasets that do not necessarily share the same biases. 
Following other well-known paradigms for identifying biases in the dataset, we first need to design or train a \emph{biased model} that employs shortcut methods to complete the task \cite{karimi2020end, sanh2020learning, utama2020towards}.

In what follows, we will first review the Product of Experts (PoE) method for bias reduction, which is based solely on the outputs of the main and biased models. Then we present our contribution to developing a novel debiasing method that takes into account not only the models' outputs but also their attribution scores.

\paragraph{Product of Experts.}
In the original Product of Experts (PoE) solution, the loss is based on combining the predictions of the main and biased model: $\sigma(f_B(\boldsymbol{x_i^b}))$ \& $\sigma(f_M(\boldsymbol{x_i}))$ \cite{training_poe_hinton, karimi2020end}. The summation of the log-softmaxes ($\sigma(.)$) of both models combine the distributions so that the main would focus less on biased examples:
\begin{equation*}
    f_C(\boldsymbol{x_i};\boldsymbol{x_i^b})=\log(\sigma(f_B(\boldsymbol{x_i^b}))) + \log(\sigma(f_M(\boldsymbol{x_i})))
\end{equation*}

The PoE loss is the cross-entropy loss over the summation shown above:
\begin{equation}
    \mathcal{L}_{\text{PoE}}(\theta_M;\theta_B)=-\log(\sigma(f_C^{y_i}(\boldsymbol{x_i};\boldsymbol{x_i^b})))
    \label{eq:poe_original}
\end{equation}

While PoE does produce promising results, training the main model only based on the biased model's output can undermine some instances that could have been helpful and non-biased. 
As stated in \citet{amirkhani2021don}, correct prediction of an instance by the biased model does not necessarily imply that the instance is biased, as the behaviour of the two models might differ.

\subsection{PoE with Saliencies}
To determine how similarly the main model and bias model behave, we must compute the attribution scores of the input tokens for both models. Therefore, after fine-tuning the biased model, we compute the saliencies $\boldsymbol{S}$ using a gradient-based approach according to the \emph{gradient$\times$input} method \cite{kindermans2016investigating}:
\begin{equation}
    \boldsymbol{S}_{i}=\left\|\frac{\partial y_{i}}{\partial \boldsymbol{h}_{i}^{0}} \odot \boldsymbol{h}_{i}^{0}\right\|_{2}
\label{eq:sals}
\end{equation}
This method is based on the gradient of the logit of the output prediction $y_{i}$ with respect to the input embeddings $\boldsymbol{h}_{i}^{0}$.
We also obtain the main model saliencies during training. By computing the saliencies for both bias and main models, it is possible to estimate the contribution of each input token in a training instance to both models' predictions. 
Therefore, we can compute the inter-model similarity---between the saliencies of the two models; for instance, using the cosine similarity metric:
\begin{equation}
    \rho = \frac{\boldsymbol{S}_\text{Main} \cdot \boldsymbol{S}_\text{Biased}}{\left\|\boldsymbol{S}_\text{Main}\right\| \left\|\boldsymbol{S}_\text{Biased}\right\|}
\end{equation}
Since the saliencies defined in \ref{eq:sals} can only have positive values, $\rho$ will always be a number between 0 and 1.
This would provide a complementary metric that shows how similar the two models behave on a specific example.
Therefore, we can modify the original PoE loss function that only incorporates the output predictions and include the inter-model similarity in the debiasing loss function:
\begin{equation}
\label{eq:new_poe}
\begin{aligned}
    \mathcal{L}_{\text{PoE+Sals}}(\theta_M;\theta_B) = \rho^*&\mathcal{L}_{\text{PoE}}(\theta_M;\theta_B) \\ 
    +\alpha(1-&\rho^*)\mathcal{L}_{\text{CE}}(\theta_M)
\end{aligned}
\end{equation}

\noindent where $\mathcal{L}_{\text{CE}}(\theta_M)$ is the cross-entropy (CE) loss on the main model. 
We also define $\rho^*$ that adjusts the inter-model similarity ($\rho$) based on the PoE loss using the following formulation:
\begin{figure*}[!t]
\centering
    \subfloat{
        \includegraphics[width=0.327\textwidth, trim=0 0 0 0, clip] 
        {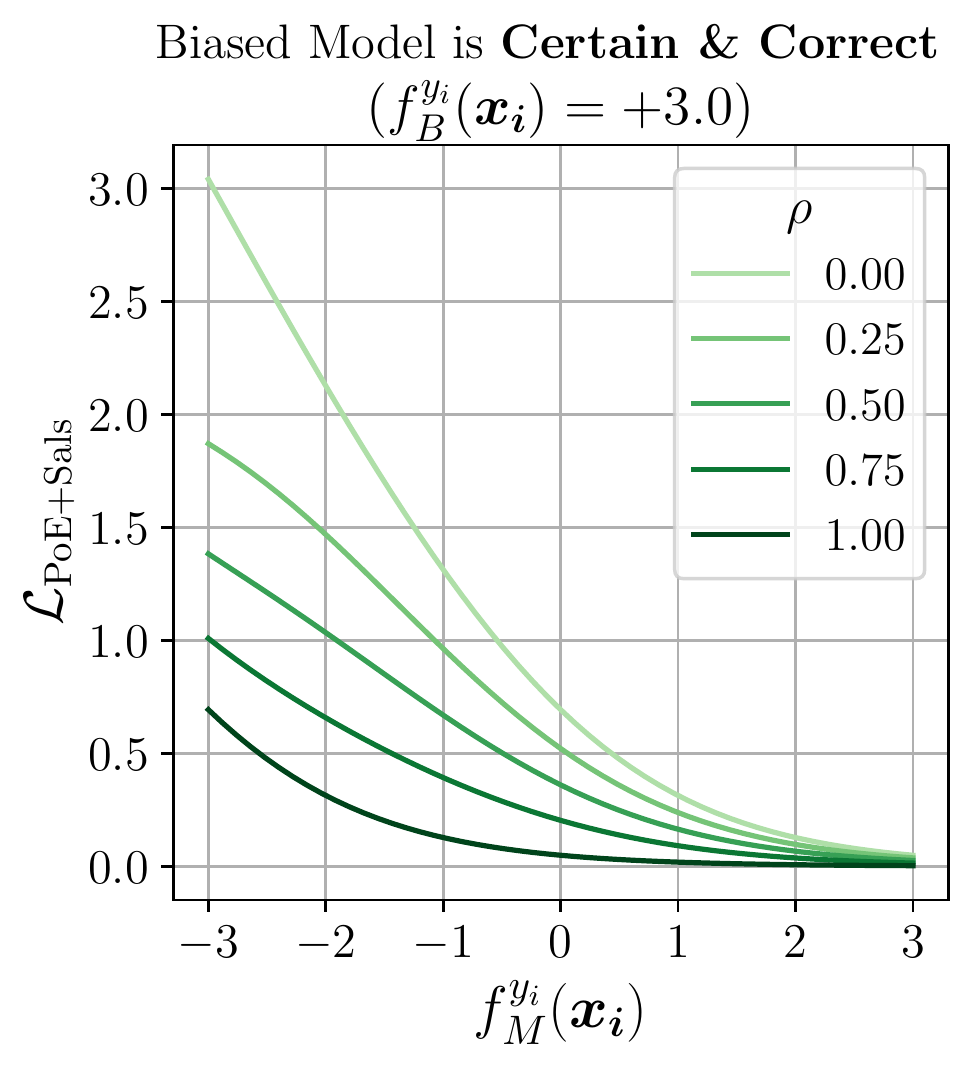}
    }
    \subfloat{
        \includegraphics[width=0.327\textwidth, trim=0 0 0 0, clip] {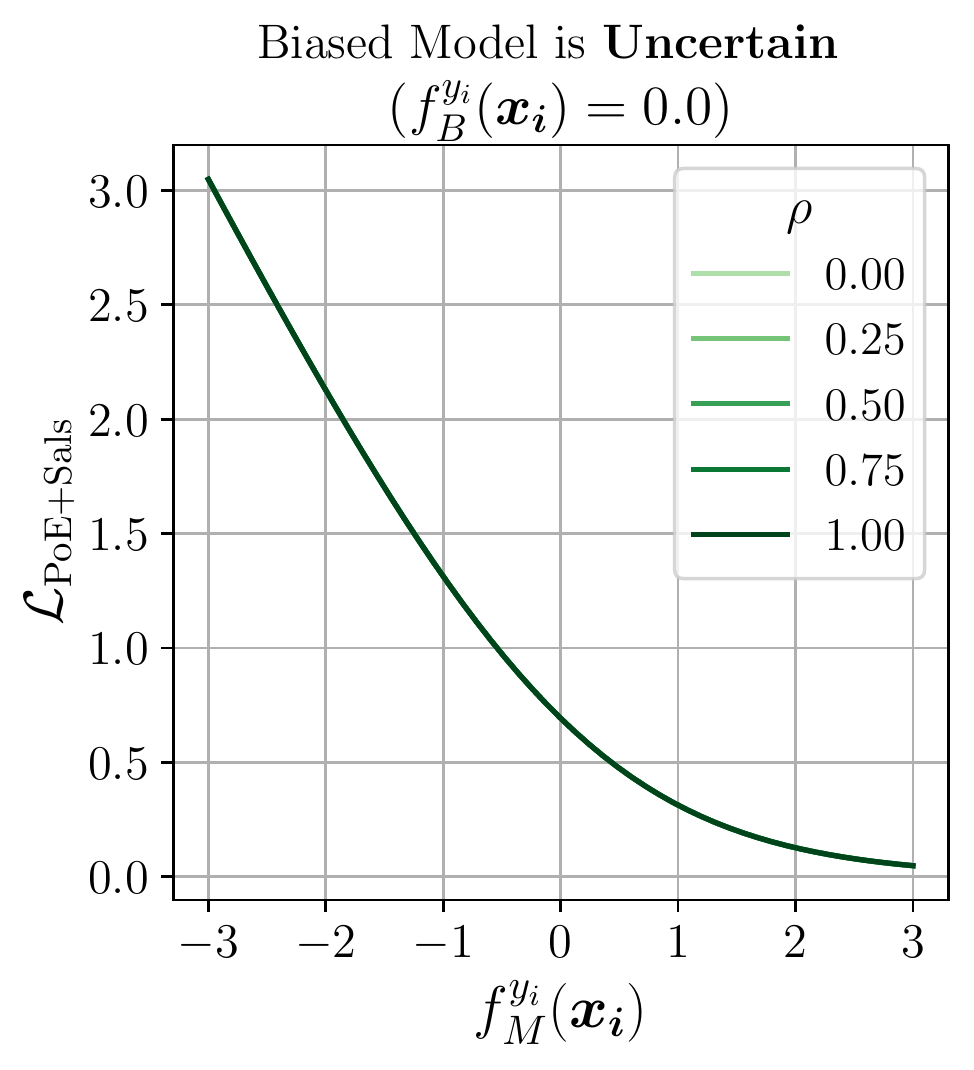}
    }
    \subfloat{
        \includegraphics[width=0.33\textwidth, trim=0 0 0 0, clip] {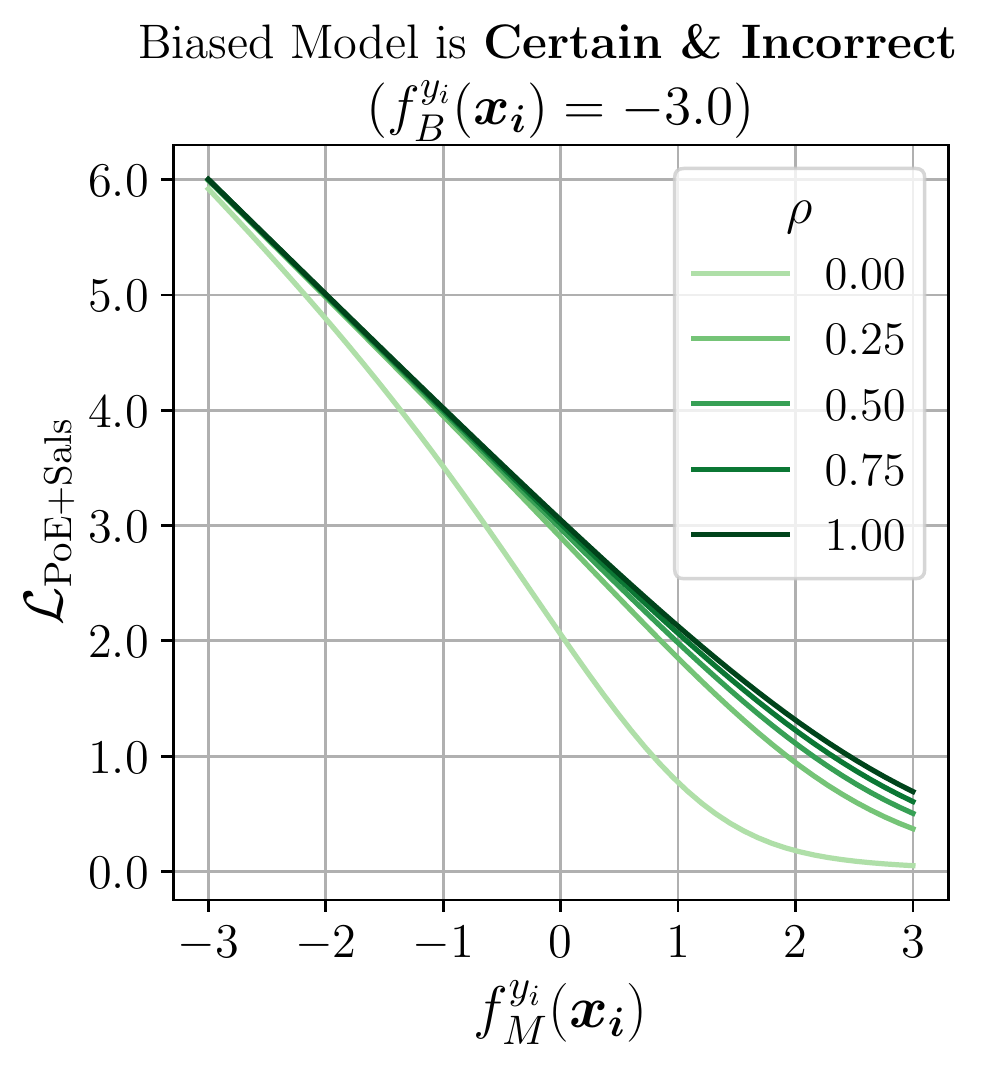}
    }
    \caption{
    {A visualization of our proposed loss function ($\mathcal{L}_{\text{PoE+Sals}}$) based on the main and biased models' prediction ($f_M^{y_i}(\boldsymbol{x_i})$ and $f_B^{y_i}(\boldsymbol{x_i})$) and the inter-model similarity ($\rho$). For a given input, when the biased model is correct and certain (left), the loss will be upweighted when the biased model differs from the main model in terms of saliency scores. However if the biased model returns a certain but incorrect prediction (right), the loss is downweighted if the main model is dissimilar and also returns a correct prediction. In an uncertain case (middle), the biased model does not affect the PoE loss \cite{sanh2020learning}. In this case, the loss is converted to a CE loss based on the output of the main model.\footref{foot:plots}
    }}
    \label{fig:poe_sals_fig_main}
\end{figure*}
\begin{equation}
\label{eq:adj_sim}
\begin{aligned}
    \rho^* &= \rho^{\exp(-\beta\mathcal{L}_{\text{PoE}}(\theta_M;\theta_B))} \\
    &= \rho^{\exp(\beta\log(\sigma(f_C^{y_i}(\boldsymbol{x_i};\boldsymbol{x_i^b}))))} \\
    &= \rho^{\sigma(f_C^{y_i}(\boldsymbol{x_i};\boldsymbol{x_i^b}))^\beta}
\end{aligned}
\end{equation}
Where $\beta$ is a positive hyperparameter for adjusting the combined prediction's ($\sigma(f_C^{y_i}(\boldsymbol{x_i};\boldsymbol{x_i^b}))$) impact.

The intuition behind this adjustment is to upweight and increase the loss for an example where the main and biased models agree on the correct label but show a different behaviour in terms of token attribution scores (a.k.a. less inter-model saliency similarity scores, $0 < \rho < 0.5$).  
As an extreme case, if the biased model correctly classifies an example with a high output score ($\sigma(f_B^{y_i}(\boldsymbol{x_i^b})) \approx 1$), the main model's prediction would be ineffective in the original PoE loss, since the PoE model output probability is as follows:
\begin{equation*}
\sigma(f_C^{y_i}(\boldsymbol{x_i};\boldsymbol{x_i^b}))=\frac{\sigma(f_B^{y_i}(\boldsymbol{x_i^b}))\sigma(f_M^{y_i}(\boldsymbol{x_i}))}{\sum_{k=1}^{Y}\sigma(f_B^{k}(\boldsymbol{x_i^b}))\sigma(f_M^{k}(\boldsymbol{x_i}))}
\label{eq:combined_f}
\end{equation*}

Therefore, in the adjusted similarity stated in Eq. \ref{eq:adj_sim}, the exponent would be approximately equal to one, which makes the adjusted similarity equal to the original cosine similarity. As a result, according to Eq. \ref{eq:new_poe}, if the models exhibit dissimilar explanations for their prediction ($\rho \approx 0$), the loss tends to be a cross-entropy loss\footnote{{Weighted with $\alpha$ as a modulating hyperparameter ($0 < \alpha$).}} than a PoE loss.
On the other hand, in case both models behave similarly on a given training example, i.e., $\rho \approx 1$, the example can be considered as one containing bias.
Thus, the PoE loss renders this example as being less impactful during training.  
The left and right plots of Figure \ref{fig:poe_sals_fig_main} respectively demonstrate\footnote{We simplify the plots by assuming the output logits of all classes except the gold label are zero for both models ($f_{B/M}^{y\neq y_i}(\boldsymbol{x_i})=0$). In addition, these figures are plotted based on $\alpha=1$ and $\beta=1$. \label{foot:plots}
} that having a higher similarity results in a loss function that is down- or up-weighted depending on the correctness of a certain biased model. Also with an uncertain biased model, the loss in Eq. \ref{eq:poe_original} converts to a CE loss which is only based on the output of the main model.

\begin{table*}[th!]
\begin{center}
\setlength{\tabcolsep}{14pt}
\scalebox{0.88}{
\begin{tabular}{l|cc|cc}
\toprule
\multicolumn{1}{c|}{\multirow{2}{*}{\textbf{Model}}} & \multicolumn{2}{c|}{\textbf{MNLI}} & \multicolumn{2}{c}{\textbf{FEVER}} \\
\multicolumn{1}{c|}{}                                & Dev. (Matched)   & HANS            & Dev.            & Sym.V1           \\ \midrule
BERT-base                                            & \textbf{84.71}$\pm$\textbf{0.21}   & 62.85$\pm$2.69  & 85.19$\pm$0.37  & 56.51$\pm$1.41   \\ \midrule
DFL$_\text{e2e}$ \cite{karimi2020end}                & 83.91$\pm$0.20   & 66.10$\pm$2.81  & 80.48$\pm$0.87  & 65.13$\pm$1.52   \\
PoE$_\text{e2e}$ \cite{karimi2020end}                & 84.10$\pm$0.19   & 63.63$\pm$1.90  & 84.43$\pm$0.87  & 64.28$\pm$1.52   \\
PoE \cite{sanh2020learning}                          & 81.10$\pm$0.41   & 68.04$\pm$1.51  & 80.08$\pm$0.94  & 62.72$\pm$2.99   \\
PoE+CE \cite{sanh2020learning}                       & 83.34$\pm$0.33   & 66.56$\pm$0.66  & 85.29$\pm$1.25  & 62.55$\pm$2.46   \\ \midrule
\textbf{PoE w. Attribution Similarity}               & 82.81$\pm$0.26   & \textbf{68.06}$\pm$\textbf{0.66}  & 85.48$\pm$1.09  & \textbf{66.97}$\pm$\textbf{2.00}   \\
    ~~~~~~~ w/o Attribution Similarity                                 & -                 & -                & \textbf{86.09}$\pm$\textbf{0.72} & 65.18$\pm$2.25  \\ \bottomrule
\end{tabular}
}
\end{center}
\caption{
The mean and standard deviation of the accuracy scores of multiple debiasing strategies applied to NLI and fact verification. The maximum values are highlighted in {bold}. It is important to note that the results of the methods that are used for comparison are not the results that were reported by the methods themselves but rather the scores that were obtained by replicating their implementation\footref{foot:fnreproduce}. 
}

\label{tab:accuracies}
\end{table*}

\section{Experiments}
In this section, we will introduce the datasets, explain the experimental setup, and then demonstrate the results of our method.
\subsection{Datasets}
The experiments were carried out on two types of common NLU classification tasks: Natural Language Inference (NLI) and Fact Verification. For NLI, we used MNLI \cite{williams-etal-2018-broad} as our in-distribution data and HANS \cite{mccoy2020right} for OOD evaluation. Note that the Matched development set is used for the ID evaluation in MNLI. In addition, because HANS has only two labels, entailment and not entailment, we consider outputs that are predicted contradictions or neutral as not entailment. For Fact Verification, we used FEVER \cite{thorne-etal-2018-fever} and FEVER Symmetric-V1 \cite{schuster-etal-2019-towards} for ID and OOD, respectively.\footnote{In both FEVER and FEVER-Symmetric, replacement tokens are used in place of parentheses and brackets \mbox{(e.g. ``]'' $\rightarrow$ ``-RSB-'')}. This causes the BERT tokenizer to split the specified tokens into multiple segments, as there are no tokens in the BERT vocabulary that correspond to the replacement tokens. Therefore, we replace these tokens with the punctuation that corresponds to them so that BERT can tokenize the inputs with less undesirable segmentation. We apply this modification to other approaches that we have evaluated and compared.}

\subsection{Setup}
In all experiments, we used \mbox{\emph{BERT-base-uncased}} \cite{devlin-etal-2019-bert} as the main model to allow a fair comparison against other debiasing methods. However, the biased model differs depending on the task. 

We used the TinyBERT model \cite{turc2019} for the NLI task, since its limited capacity makes it extremely susceptible to biased features in training examples \cite{sanh2020learning}.
For fact verification, a full 12-layer BERT-base model similar to the main model is fine-tuned using only the claim sentences from the training data. In both biased models, we compute and save their prediction logits and attribution scores across the entire training dataset so that they can be used as a frozen model during training. However, since the fact verification biased model is only trained on the claim sentences, the attribution scores are only calculated for the claim portion. Therefore, the similarities computed during the training procedure are limited to the claim segment alone.

For the generic fine-tuning hyperparameters, we adopted \citet{sanh2020learning} recommendations: 3 epochs of training, a batch size of 32, an Adam optimizer \cite{adamKingma} with warmup and linear decay in its learning rate schedule, and a peak learning rate of 3e-5 or 2e-5 for MNLI or FEVER, respectively. But as for the specific hyperparameters in our approach, using sweeping over $\alpha \in \{0.01, 0.1, 0.2, 0.3, 0.5, 1.0\}$ and $\beta \in \{0.1, 0.3, 0.5, 1.0\}$, we set $\alpha=1.0$, $\beta=1.0$ for MNLI and $\alpha=0.3$, $\beta=0.1$ for training on FEVER.

All experiments were implemented using the HuggingFace Transformers library \cite{wolf-etal-2020-transformers} and performed on an RTX 3070 GPU machine. The results are the average of six runs with different seeds.
\subsection{Results}
The results of various debiasing techniques applied to the previously mentioned benchmarks are shown in Table Table \ref{tab:accuracies}.
The baseline is fine-tuning the backbone model (BERT-base-uncased) with the commonly used cross-entropy loss, which provides high ID performance but lacks OOD. We also include four approaches from two different studies for additional comparison\footnote{We could have included the compared methods' results from their respective papers as well as results from other approaches. However, in our empirical results we observed significant discrepancies between the results obtained from their source code and those reported. In this paper, we intend to report only the reproduced results, which is why the number of compared methods is limited. \label{foot:fnreproduce}}. Two End-to-End solutions from \citet{karimi2020end}, one utilizing Debiased Focal Loss (DFL) and the other employing PoE. The PoE and PoE+CE (PoE with a static weighted CE loss added) methods from \citet{sanh2020learning} are similar to our approach of having a biased model that is frozen, but they only rely on the predictions of the biased model and not the inter-model similarities.

It can be observed that the OOD performance of our method outperforms those of other approaches, without or with minimal loss of ID accuracy. On FEVER, our strategy improves SymV1 by {nearly 2\% while also slightly improving the ID performance}. In MNLI, our strategy improves HANS to achieve high scores comparable to \citet{sanh2020learning} PoE-only solution. However, the ID performance of \citet{sanh2020learning} PoE-only falls short. While maintaining the same level of OOD performance, our solution achieves a minimum MNLI-m dev score that is acceptable.

As an ablation study, we also trained the procedure without using similarity values, resulting in a PoE+CE solution. Since we used a different biased model in FEVER than \citet{sanh2020learning}, we reported its results using the claim-only biased model configuration. However, we omitted MNLI results from our report because they are identical to \citet{sanh2020learning} PoE+CE configuration. Having a claim-only biased model for FEVER rather than a TinyBERT model yields a substantial increase in OOD, as can be seen in the results. As a result, we can observe that even though adding a weighted CE to a PoE loss yields improvements in terms of ID performance in particular, the similarities could push even further and also improve OOD performance.

Another observation is the large standard deviations in OOD accuracy of all approaches, which is why we chose to execute 6 runs as also suggested by \citet{sanh2020learning}. Even so, it should be noted that our method still has a relatively low variance in HANS.

\section{Conclusions}
In this paper, we introduced a strategy for improving OOD performance by incorporating the similarity values between the token attribution scores of main and biased models into a Product of Experts (PoE) loss function. 
The gist of this approach is that it takes into account the decision-making process in addition to the output predictions of the main and biased models (which are often taken as the primary signal). 
By comparing our method to multiple recent debiasing methods on two widely-used NLU tasks, we demonstrated that our method improves performance on out-of-distribution data while preserving performance on in-distribution data. 
Future work could include exploring other loss formulations with Attribution Similarities that may produce better results. 
Additionally, it may be beneficial to investigate methods to reduce the variability of results, which is present in the majority of debiasing solutions.

\section*{Limitations}
Due to the high variance in the accuracy scores, the main limitation of this approach (as well as the majority of other debiasing approaches) is the large number of seeds used to tune the hyperparameters. As a result, any type of tuning necessitates a multi-seed run, which requires multiple GPUs or many hours of training. Because of this constraint, we omitted working on larger scaled models or other types of PLMs.


\bibliography{custom}
\bibliographystyle{acl_natbib}





\end{document}